\documentclass[10pt,journal,compsoc]{IEEEtran}

\usepackage{booktabs}
\usepackage{multirow}

\ifCLASSOPTIONcompsoc
  \usepackage[nocompress]{cite}
\else
  \usepackage{cite}
\fi

\usepackage{graphicx}

\usepackage{amsmath}

\usepackage{algorithmic}

\begin{document}
%
\title{No-audio speaking status detection in crowded settings via visual pose-based filtering and wearable acceleration}
%
%
%
%
 
\author{Jose~Vargas-Quiros,
		Laura~Cabrera-Quiros,
        and~Hayley~Hung
\IEEEcompsocitemizethanks{\IEEEcompsocthanksitem J. Vargas and H. Hung were with the Department of Intelligent Systems at TU Delft, The Netherlands.\protect\\
E-mail: {j.d.vargasquiros, h.hung}@tudelft.nl
\IEEEcompsocthanksitem L. Cabrera Quiros was with Escuela de Ingenieria Electronica at the Instituto Tecnologico de Costa Rica, Costa Rica.}
\thanks{Manuscript received April 19, 2005; revised August 26, 2015.}}

%
%


\IEEEtitleabstractindextext{%
\begin{abstract}
Recognizing who is speaking in a crowded scene is a key challenge towards the understanding of the social interactions going on within. Detecting speaking status from body movement alone opens the door for the analysis of social scenes in which personal audio is not obtainable. Video and wearable sensors make it possible recognize speaking in an unobtrusive, privacy-preserving way. When considering the video modality, in action recognition problems, a bounding box is traditionally used to localize and segment out the target subject, to then recognize the action taking place within it. However, cross-contamination, occlusion, and the articulated nature of the human body, make this approach challenging in a crowded scene. Here, we leverage articulated body poses for subject localization and in the subsequent speech detection stage. We show that the selection of local features around pose keypoints has a positive effect on generalization performance while also significantly reducing the number of local features considered, making for a more efficient method. Using two in-the-wild datasets with different viewpoints of subjects, we investigate the role of cross-contamination in this effect. We additionally make use of acceleration measured through wearable sensors for the same task, and present a multimodal approach combining both methods.
\end{abstract}

}

\maketitle

\IEEEdisplaynontitleabstractindextext
\IEEEpeerreviewmaketitle

\IEEEraisesectionheading{\section{Introduction}\label{sec:introduction}}

\begin{figure}[!t]
\includegraphics[width=\columnwidth]{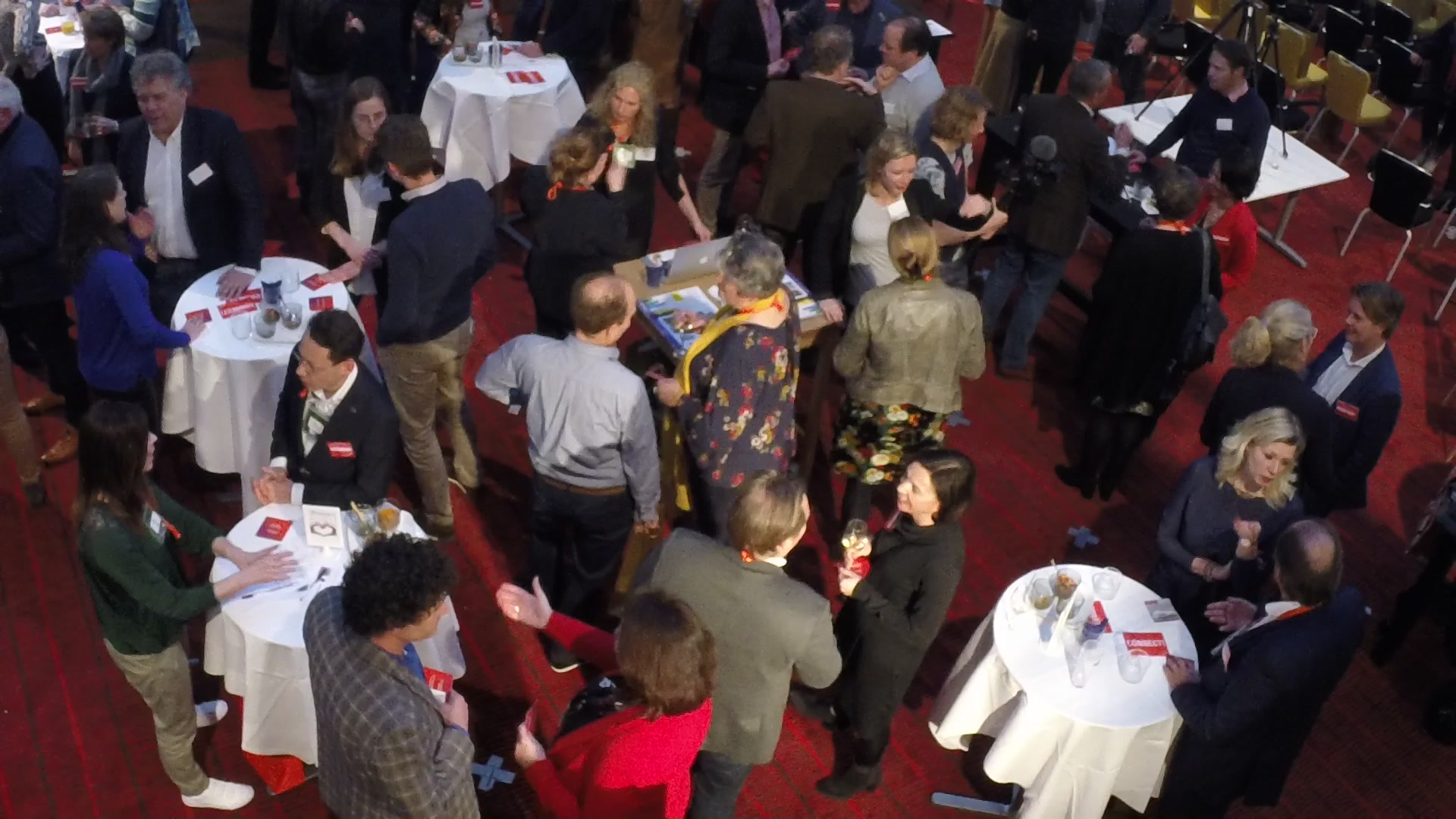}
\caption{Example of a free-standing crowded scene from our data.}
\label{fig:cameras}
\end{figure}

\IEEEPARstart{D}{etection} of speaking activity in free-standing social settings is a core necessity in building systems capable of detecting and understanding the social interactions that take place in a scene. The analysis of a complex conversational scene where dozens of people stand, walk, form groups and converse freely (see Fig. \ref{fig:cameras}) is of interest in fields such as computational social science and social signal processing. Speaking status is key because of its utility in downstream tasks, where it can be used, for example, in the quantification of individual and group measures of conversation quality like involvement \cite{Oertel2011}, satisfaction \cite{Lai2018} or affect \cite{Lai2013}, and in the forecasting of future events like speaking, gesturing and changes in position and orientation \cite{Joo2019}.

This goal requires systems capable of efficient speaking status detection, due to the large number of people in the scene and the extent of natural human interactions. Although the audio modality is the obvious choice for the measurement of speaking status, this modality is especially hard to acquire in a crowded conversational scene. It is possible to obtain high-quality speaking status from personal head-mounted and directional microphones, but such equipment is hard to scale for the study of large crowds and introduces privacy concerns due to the necessary access to the content of conversations. The use of ubiquitous technology via wearables like sociometric badges \cite{Alameda-Pineda2016} and mobile phone sensing solves the scaling issue and may alleviate privacy issues through the sensing of lower-fidelity signals that obscure the content of conversations \cite{Lederman2017}. However, the problems introduced by the large number of sound sources in a crowd have not been addressed in source separation methods and pose severe challenges when the audio is captured via the omnidirectional microphones common in most mobile and wearable devices \cite{Alameda-Pineda2016}. In fact, most datasets created for the study of free-standing crowds do not contain audio \cite{Cabrera-Quiros2018a}.

The possibility of detecting speaking from body movement alone, without access to the audio, offers a privacy sensitive solution to these problems. It has long been observed that hand and head gestures frequently synchronize with speech \cite{Lascarides2009, McNeill1994} while being salient cues with similar motion characteristics across people. The usual video modality offers convenient way to observe these gestures but is affected by factors like occlusion and cross-contamination. Wearable accelerometers are capable of capturing more subtle body movements in 3D space while being invariant to the same factors. Previous work has shown the utility of these signals in detection of different social actions \cite{Hung2014, Hung2013a, Gedik2017, Cabrera-Quiros2018b, Vargas2019a}. We are therefore interested in the combination of these two modalities for the detection of speech activity.

Regarding the visual detection of the speaking status of a target subject in the scene, the related challenges of cross-contamination and occlusion are made worse when the person is localized using a bounding box \cite{Cabrera-Quiros2018a} due to the difficulty of accurately segmenting a person's body. We consider cross-contamination to include any case where the bodies of other people in the scene are visible within the bounding box, or area considered by the recognition system and occlusion when parts of the body of the target subject are occluded, possibly by their own body \cite{Cabrera-Quiros2018a}. 

Given recent significant advances in visual pose estimation \cite{Cao2017, Xiu2019, Li2019a} it is natural to think of the application of these methods to alleviate these issues. Accurate poses allow for more precise localization and filtering of the information that is input to the recognition stage. While existing works have addressed pose-based action recognition \cite{Yan2019, Choutas2018}, including feature filtering \cite{Konstantinos2017}, results have been noted to be dependent on the particular actions being performed \cite{Pishchulin2014}. These results, however, have been obtained in action recognition datasets with little cross-contamination and less subtle, more distinct actions.

In this work, we propose a multimodal method combining visual pose-based feature filtering and acceleration-based detection applied to the problem of speaking status detection in a crowded room. Our contributions are the following:

\begin{enumerate}
\item We collected an in-the-wild dataset with video and individual audio and body-worn accelerometer readings, in a crowded setting. In contrast with previous work relying on visual annotations of speaking status \cite{Cabrera-Quiros2018a}, we obtained our ground truth automatically from high-quality voice recordings.

\item We propose a method for speaking status detection that selects trajectories around pose keypoints. We made use of body poses extracted automatically via a state-of-the-art method. We show that focusing on upper body keypoints, and head and hand keypoints in particular increases speaking status detection performance while decreasing the number of considered features, making for a more efficient method. Via evaluation on two datasets with different viewpoints and crowded-ness levels, we found that the relative importance of information from the head and from the hands keypoints was dataset dependent, but with head trajectories having a consistently high importance.

\item We proposed a multimodal method combining the pose-based method (motion and appearance streams) with an acceleration stream making use of body acceleration signals, which are not affected by cross-contamination. We evaluated the multimodal method against our collected dataset and show that it improves over single-modality methods.
\end{enumerate}

\section{Related work}

\subsection{Visual Detection of Actions}
The action recognition field in computer vision has long studied the problem of recognizing human actions in videos. Traditional approaches include the extraction of dense trajectories \cite{Wang2011}, including their spatio-temporal descriptors, followed by an encoding method which transforms the trajectories from a video into a single high-dimensional video-level representation \cite{Peng2016}. Fisher Vectors is the most prevalent and well-studied of such encoding methods due to its superior performance \cite{Oneata,He2005,Sanchez2013}. Improved dense trajectories \cite{Wang2013} correct for camera motion and use bounding box human detections to filter out surrounding trajectories.

More recent approaches make use of the power of convolutional neural networks (CNNs). Two-stream networks feed static frames from the video into an appearance stream and optical flow computed between frames into a motion stream, the scores of which are fused together for classification, effectively creating a separation between static and motion information \cite{Desk2006}. Recently, 3D CNNs \cite{Ji2013, Tran2015} have gotten increased attention. Pre-training techniques using large datasets have demonstrated performance improvements over 2D CNNs and traditional approaches \cite{Zisserman2018}. The more recent (2+1)D CNNs cover a middle ground by factorizing 3D convolutional filters into 2D spatial and 1D temporal convolutions \cite{Tran2018a}, with however similar performance.

Although both 3D and (2+1)D CNNs deliver state of the art results in most action recognition benchmarks, they are also notorious for requiring training in large-scale datasets to achieve such performance. While the use of pre-trained models is a possibility for smaller datasets, the significant domain shifts between the kind of actions and viewpoints considered in large-scale datasets and a target dataset may make a model trained from scratch preferable. Despite the improvements in CNN efficiency, improved dense trajectories has featured close to state-of-the-art performance in the recent Charades dataset of everyday human activities \cite{Sigurdsson2016, Sigurdsson2017} while remaining competitive in traditional action recognition datasets among methods with no pre-training \cite{Zisserman2018}.

 
The development of faster and more accurate pose estimation and tracking systems capable of detecting the poses in a video at close to real time \cite{Cao2017, Li2019a, Fang2017, Xiu2019} has resulted in increased interest in pose-based action recognition systems. Previous work has assigned trajectories to joints, to later learn a separate bag-of-words per joint \cite{Konstantinos2017}. However, the method is designed and tested with frontal views of a single actor, and therefore does not perform any filtering. While learning separate representation per joint works in such cases, it is impractical in a crowded setting where most joints are very frequently occluded.

Previous work \cite{Cheron2015} extracts RGB and optical flow patches around each pose keypoint and feed them into one appearance and one motion CNN, effectively a two-stream network per keypoint. Similarly, Pishchulin experimented with multiple ways of combining ground truth pose and dense trajectory information, including filtering trajectories using square regions around pose keypoints \cite{Pishchulin2014}. Experiments were performed on a dataset with over 800 human activities. Performance improvements were highly dependent on the specific action being detected. No further details are given about the way trajectories were filtered.


\begin{figure*}[!t]
	\centering
	\includegraphics[width=0.7\textwidth]{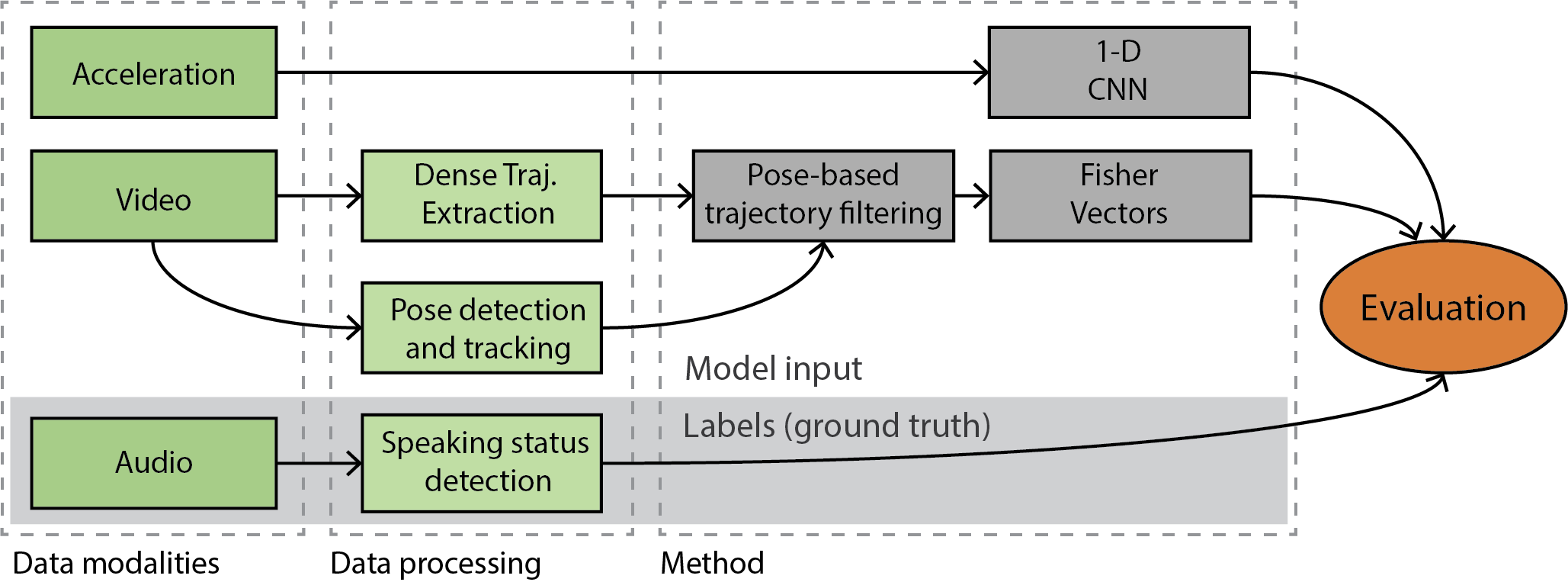}
	\caption{Overview of our work, from data collection to speaking status detection.}
	\label{fig:diagram}
\end{figure*}

\subsection{No-audio Speech Detection}

Despite the significant amount of work on generic action recognition and localization, interest for the problem of detecting the speaker in a conversation has been limited.

The most closely related vision works have addressed the problem of detecting speaking status from body movement information alone \cite{Shahid2019}, introducing the dataset called RealVAD for this purpose \cite{Beyan2020}. This dataset consists of a single-camera frontal recording of a panel discussion where 9 subjects take turns speaking. Even though this dataset does not contain a free conversation, it does capture a real event where subject's body movement is clearly visible. This dataset also has some cross-contamination, due to views of subjects overlapping with each other, albeit much less and less variable than in our free-standing conversation setting with multiple angles. The methods presented make use of domain adaptation to adapt CNN-extracted features of one speaker to another, using bounding boxes to localize and crop out the subject regions.

Related problems have also received some attention. One of them is the problem of speaker naming in movies \cite{Hu2015, Roth2019}, where the goal is to localize and identify the speaking character in a movie or video. However, an important difference is that in movie naming the algorithm has access to the audio, and movie scenes tend to have clear frontal views of speakers, making it possible to rely on face detection and tracking \cite{Hu2015}.


On the other hand, acceleration readings have been used successfully for the assessment of human actions, more commonly for the recognition of daily activities like walking or running, but also for social actions including speech. Wang surveyed state of the art deep learning approaches for sensor-based activity recognition, including accelerometers \cite{Wang2018a}. A study using chest-worn accelerometers established their ability to recognize actions like gesturing, laughing, and speaking in a free-standing social setting similar to ours \cite{Hung2013a}. The best results were found for the recognition of speaking, with more recent work improving on the methods used \cite{Gedik2017}.

\subsection{Multimodal speech detection}
Work on multimodal speaking gesture detection \cite{Cabrera-Quiros2018b} is the most closely related to ours in method, presenting a multimodal method for gesture and speaking status detection in a crowded scene. The video-based method is inspired on Multiple Instance Learning for trajectory aggregation and classification, and combined with an acceleration modality via late fusion. However, this study relies on human-annotated bounding boxes and speaking status annotations, and does not focus on solving the issue of cross-contamination.

Multimodal approaches in similar settings have found that a single accelerometer offers performance competitive with trajectory-based video recognition of speaking status \cite{Cabrera-Quiros2018b, Vargas2019a}. We believe however that these modalities in many cases compliment each other, and that a method will benefit from access to both, as these previous work has also shown.

\subsection{Summary}

The main challenges in detecting speaking status from body movement in a crowded scene using existing methods are the subtlety (low intensity and visual saliency) of the movements involved (in comparison with the movements present in most action recognition datasets) and cross-contamination from other people in the scene. A smaller set of previous works looking specifically at speaking status detection without audio \cite{Shahid2019, Beyan2020, Cabrera-Quiros2018b, Gedik2017} have addressed the first of these problems. However, to the best of our knowledge the second problem remains un-addressed. In this paper we attempt to understand and tackle this problem via existing action recognition methods using accelerometers and video data.

\section{Datasets}

We made use of two datasets with available speaking status annotations to validate our approach. The two datasets differ in having significantly distinct views of the subjects and in the setting in which they were collected. Both datasets, however, share the issue of cross-contamination in the video modality. The first dataset, RealVAD, was published as part of a paper analyzing no-audio speaking status detection \cite{Beyan2020} in a panel session, where subjects take long speaking turns and are recorded from the front with little cross-contamination. We collected a more challenging second dataset by recording an in-the-wild mingling event. This dataset has free interaction between 43 recorded participants, with frequent turn-taking and heavy cross-contamination. 

\subsection{RealVAD dataset}

The RealVAD dataset was presented in \cite{Beyan2020}. It was collected during a panel session lasting approximately 83 minutes, where eight panelists took turns addressing the audience, and has therefore infrequent turn changes. All panelists are captured via a single frontal camera with only mild cross-contamination between them. Although this dataset has relatively little cross-contamination, we are also interested in testing our approach on a dataset where body parts are more consistently visible, to better understand their relative importance as indicators of speaking. Furthermore, frontal shots like those in RealVAD are common in video datasets, and we wish to understand how the performance of the method changes under such condition.

\subsection{Free-standing dataset}

Investigating speaking status in the wild requires the collection of a dataset in which social interaction occurs with as little intervention as possible. To this end, we collected an additional dataset, which, in contrast to RealVAD, contains standing participants in free interaction. As in RealVAD, subjects wore microphones used to obtain their speaking status.

The dataset was collected during a special event organized by a business networking group. Most participants in the event meet regularly and many but not all of them knew each other. Participants were informed beforehand that this particular meeting would be recorded. As they arrived to the event, they were asked for consent after being further informed about the data collection. They were free to choose which sensors to wear (microphone, accelerometer, or both) or to not participate in the data collection. All participants were informed about a clearly-delimited video zone where they would be recorded by our video cameras. This process was approved by the ethics board of the university beforehand.

Of about 100 attendees to the event, 43 consented to wearing a microphones or accelerometer sensor. Of them, 20 were male and 23 female. Most of the participants interacted within the video zone. 

During the event, most of the interaction consisted of free-standing conversation. Participants were free to move around and talk as they pleased. Because many participants were acquainted with each other and this was a special event commemorating an anniversary of their organization, conversations were mostly friendly and sporadic.

\subsubsection{Sensor setup}

We collected data using the following sensors:

\begin{itemize}
	\item A custom-made wearable accelerometer sensor that was hung around the neck and rested on the chest like a smart ID badge.
	\item Lavalier microphones attached to the face to record speaking activity, used for obtaining the ground truth labels. Microphones were attached to a Sennheiser SK2000 transmitter. Audio was recorded at 48kHz.
	\item 12 overhead cameras and four side-elevated cameras were placed above and in the corners of a video zone. In this work we only make use of the side elevated cameras.
\end{itemize}


Because many participants chose to only wear one of the sensors or to not enter the video zone, and because of the malfunction of some of our wearable devices, not all modalities were available for all participants. Table \ref{tab:modalities} shows the dataset statistics where we see that 17 out of a possible 43 participants had data from both modalities available to them. This highlights the challenges associated with capturing in-the-wild data where participants can choose what data to provide. However, we can be more confident of the realism of the data compared to more controlled settings. 

Our recordings included segments when the participants were expected to listen to a speaker or listen to a performance. We used the video modality to manually find and eliminate such segments, as they deviate from our setting of interest. 

\subsubsection{Automatic speaking status annotation}

Due to the availability of high-quality audio recordings of our subjects, we first explored the automatic annotation of speaking activity via voice activity detection (VAD) algorithms. We first investigated the feasibility of using pre-trained neural models, trained on both the AMI dataset, and the more diverse and challenging DIHARD dataset \cite{Ryant2019}, through the \textit{pyannote.audio} package \cite{Bredin2020, Lavechin2019}. However, we found these methods to be too sensitive for our use case, detecting most background speech with high confidence. 

\begin{figure}
	\includegraphics[width=\columnwidth]{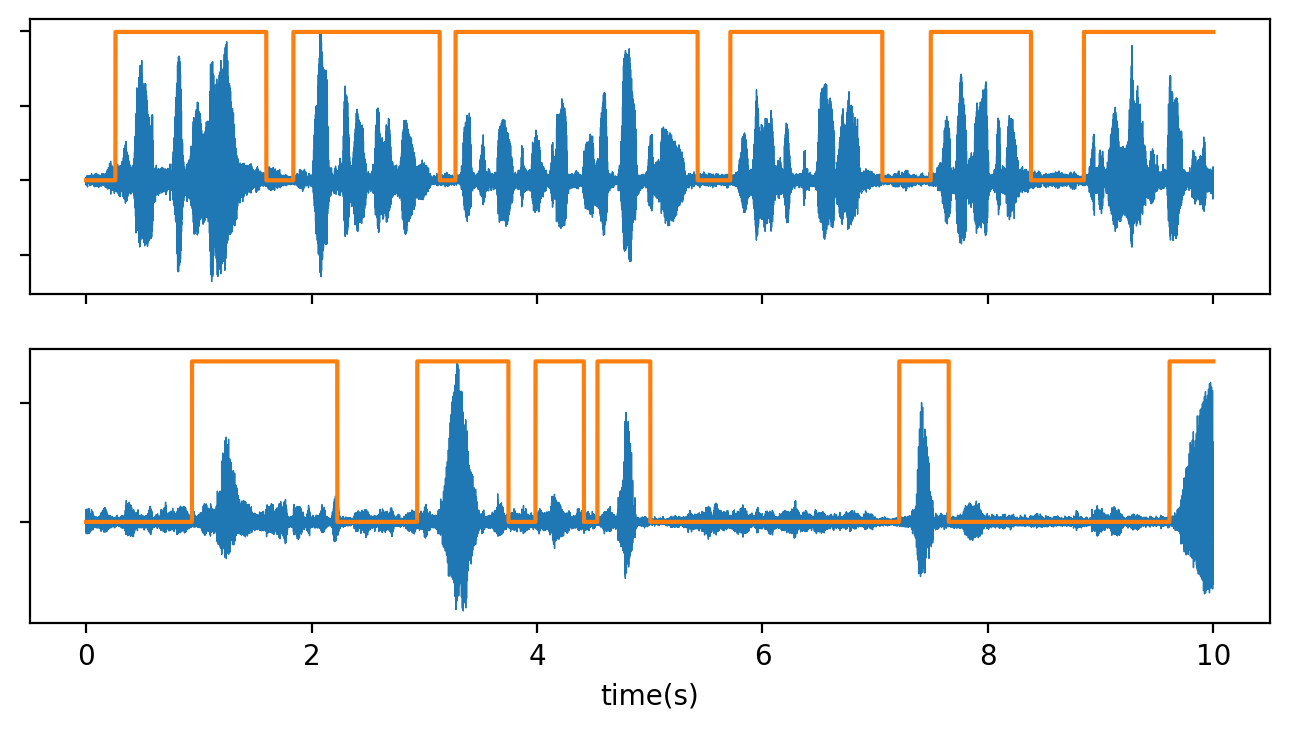}
	\caption{Example of our speaking status detections showing the audio waveform and VAD binary predictions for two data segments. In the first case the speaker takes few pauses, in the second it is back-channeling to its interlocutor.}
	\label{fig:rvad}
\end{figure}

Due to the closeness to the mouth of our head-worn microphones, there was a significant difference in energy between speech of the speaker and background speech. The presence of background noise, however, poses a challenge for VAD. We therefore relied on the rVAD method for robust voice activity detection \cite{Tan2020}, which relies on pitch detection, applies several de-noising passes, and directly takes into account signal energy differences in segmentation of the speech signal. We used the full version of the rVAD detector to produce binary speaking status outputs for our participant recordings at 100Hz. We found this method to reliably segment speaker voice activity from background noises. Fig. \ref{fig:rvad} shows an example of two output segmentations.

\begin{table}
	\caption{Statistics of our speaking status dataset. The number of positive examples appears in parentheses.}
	\label{tab:modalities}
	\centering
	\begin{tabular}{ccll}
		\toprule
		Modalities & Participants & Num. Examples & Hours\\
		\midrule
		Video & 24 & 18639 (10635) & 15.53 \\
		Video \& Acceleration & 17 & 12309 (7695) &  10.26\\
		\bottomrule
	\end{tabular}
\end{table}



\section{Approach}

In this section we justify and detail our decisions on the method using both the video and acceleration modalities. An overview of our approach, including data collection and processing is shown in Fig. \ref{fig:diagram}.

\subsection{Voice activity detection from video}
We start by describing the processing done on the video modality, including the process used to extract pose tracks from videos of human interaction.

\subsubsection{Pose estimation}\label{sec:pose_estimation}

Accurate pose detection is central to our approach, which relies on it for local feature selection. We considered the two most well-known and maintained pose estimation methods: Openpose \cite{Cao2017}, a bottom-up approach based on part affinity fields; and AlphaPose, a top-down approach \cite{Fang2017}. Both methods achieve competitive results in well-known pose estimation benchmarks like MPII Human Pose \cite{Andriluka2014} and MS COCO Keypoints. Our tests using both methods on a small representative set of data revealed that both methods were able to detect most body keypoints consistently across frames, with some keypoints being more reliable than others (with upper body keypoints being more reliable in general). We decided to use Openpose (BODY25 model) due to its faster running time, independent of the number of people in the frame.

\subsubsection{Pose tracking}\label{sec:pose_tracking}

Due to the fact that most pose estimation algorithms, including OpenPose, work independently on individual frames, we required a way to associate poses across frames. While this problem has been investigated in previous work, we found the existing PoseFlow \cite{Xiu2019} to be too computationally expensive for our use case, due to the large number of people in the scene. We therefore implemented a semi-automatic method to obtain tracks from individual frame detections. We opted for a computationally lighter method based on the observation that the chest keypoint was reliably detected and localized across frames. 

Our goal is to create pose tracks by associating the chest keypoint across frames. Concretely, for each frame $n$ of the video, the pose detector outputs a set of poses given by $Q_n = \{P_{n,m} \mid m = 1, \dots, M_n\}$, where $M_n$ is the number of people detected in frame $n$ and $P_{n,m} = \{\pmb{p}_{n,m,j} \mid j = 1, \dots, J\}\text{; }\pmb{p}_{n,m,j} = (p_{n,m,j}^{\hat x}, p_{n,m,j}^{\hat y})$ is a vector of $J$ 2D-keypoints representing a skeleton. A pose track $U$ is a sequence of consecutive poses given by $U = \{P_{i_U}, \dotsc, P_{f_U}\}$, starting at frame $i_U$ and ending at frame $f_U$.

With the same goal of a fast method, we decided for a step-wise approach, where the goal is to match poses in $Q_n$ from frame $n$ to tracks consisting of poses from all frames up to $n-1$. Specifically, we solve the assignment problem between two sets of poses: $Q_n$ and $\{P_{f_U} | U \in T_n \text{ and } n - f_J < R_{th}\}$, for $n = 1,2,...,N$. This is repeated in order for $n=1,\dotsc,N$. In other words, we compare poses in $Q_n$ with the head of existing tracks whose last pose is not older than $R_{th}$ frames, where $R_{th}$ is an integer parameter.
 
The assignment problem is defined by a distance calculation. We define the distance between two poses as the Euclidean distance between their chest keypoints $D(P_A, P_B) = ||\pmb P_A^{chest} - \pmb P_B^{chest}||$. We solve the assignment problem via the Hungarian algorithm. We add a maximum distance threshold $D_{th}$ for assignment, such that if $D(P_A, P_B) > D_{th}$, then $P_A$ and $P_B$ cannot be assigned to each other. Assigned keypoints are added to the corresponding track and unassigned keypoints are assigned to a new track. When a new pose in frame $n$ is matched to a pose in a frame $m \neq n-1$ (not the immediately preceding frame), we impute the keypoints via linear interpolation to maintain the continuity of the track. 

Parameters $R_{th}$ and $D_{th}$ were set based on experiments on a subset of the tracks. This approach resulted in high quality tracks, with only few person switches due to subjects walking in front of one another. We found the one-second threshold to work well in eliminating issues of consistency across frames without introducing significant errors.

Because our goal was to obtain high quality tracks to be able to reliably test our recognition method (see next sections), we manually inspected the dataset for track switches and corrected them by splitting the tracks. We further assigned tracks to the corresponding ID of the participant to be able to associate with the personal acceleration readings. 


\subsubsection{Dense trajectories}
\label{sec:dts}

Research in psychology and social signal processing has shown that body gestures, and especially hand gestures are closely synchronized with speech \cite{Esposito2011, Pouw2019}. Because of the difficulty of accessing facial information in our setting, our method aims to capture primarily such gestures and overall body movement. For our video-based detection method, we rely on dense trajectories \cite{Wang2011} due to their ability to track such salient movements. Additionally, the relatively small size and non-standard viewpoint of our dataset make a more simple method trained from scratch preferable to a method based on pre-trained CNNs. Given that we don't have camera motion, we make use of the original dense trajectories \cite{Wang2011}, and not improved trajectories \cite{Wang2013}.

\begin{figure*}[!t]
	\centering
	\includegraphics[width=0.7\textwidth]{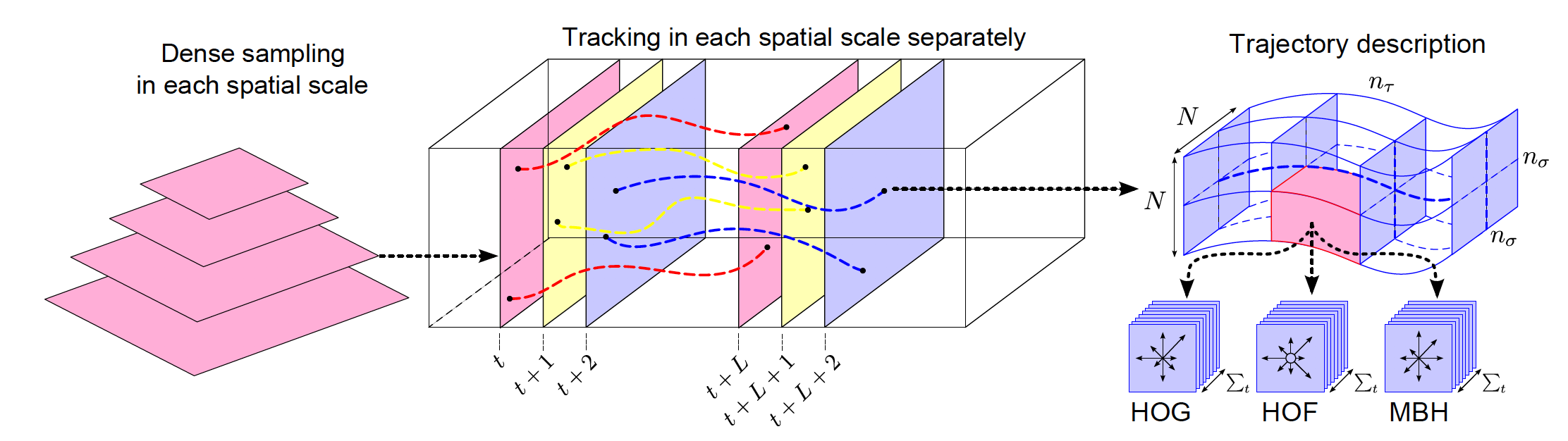}
	\caption{Dense trajectories are created by sampling interest points from multiple scales. These are then tracked by following the optical flow fields over $L$ frames. Finally, features are extracted that describe the volume around the trajectory. \cite{Wang2011}}
	\label{fig:dts}
\end{figure*}

Dense trajectories \cite{Wang2011} were proposed in action recognition literature for the classification of short videos labeled with the action being performed in them. They are extracted by a process of sampling feature points and subsequently tracking them for the following frames using optical flow. Feature points are sampled on a grid spaced by $W=5$ pixels in 8 spatial scales spaced by a factor of $1/\sqrt{2}$. Points are then tracked using a dense optical flow field. This is done for $L=15$ frames in order to avoid drift from longer trajectories. The spatio-temporal volume in a neighborhood of size $N=32$ pixels around the trajectory is then described using Histograms of Gradients (HOG), Histograms of Optical Flow (HOF) and Motion Boundary Histograms (MBF) features extracted from cells dividing the volume in a grid of size $n_{\lambda} \times n_{\lambda} \times n_{\tau} = 2 \times 2 \times 3$ (see Fig. \ref{fig:dts}). While HOG features are predominantly visual, HOF features capture more temporal information. Motion boundary histograms capture both visual and temporal information.

The mentioned dense trajectory parameters: length $L$, step size $W$, neighborhood size $n_{\lambda}$, $n_{\tau}$ were set to their default values, which are replicated in most previous action recognition literature using trajectories and shown to be close to optimal on different datasets \cite{Wang2011}. The final dense trajectory descriptor is the concatenation of the trajectory (size 30), HOG (size 96), HOF (size 108), and MBH (size 192) vectors for a total of $D=426$ dimensions describing a video segment.

\subsubsection{Pose-based filtering of dense trajectories}
\label{sec:pose-aligned}

The goal of our filtering approach is to reduce the effect of spurious trajectories due to cross-contamination by people other than the target subject. At the same time, we do not attempt to precisely segment the subject. While having precise segmentations would be ideal, person instance segmentation methods \cite{Guler2018} add significant computational complexity and in preliminary experiments had clearly worse performance in detecting joints in our dataset.

\begin{figure*}[!t]
	\centering
	\includegraphics[width=0.8\textwidth]{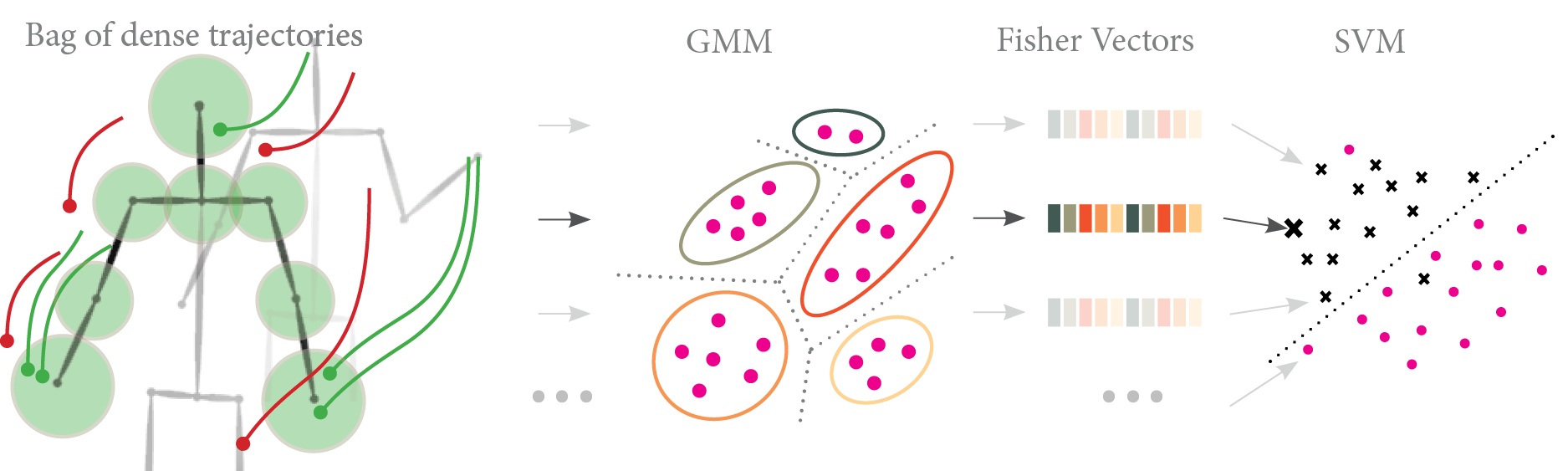}
	\caption{Our video-based approach selects trajectories around pose keypoints.}
	\label{fig:pose_fvs}
\end{figure*}

Due to the frequent occlusion of lower-body keypoints as a result of the crowd density and view angle of the camera, we only consider $J = 8$ upper body keypoints (head, neck, shoulders, elbows and wrists as sown in Fig. \ref{fig:pose_fvs}). The single head keypoint was obtained by averaging the eyes, nose and ears keypoints output by the pose estimator. Our trajectory filtering method picks trajectories starting within a distance (in the image frame) of these 8 keypoints, and filters out the rest. We use the first point of the trajectories because the points in a dense trajectory accumulate errors due to the use of optical flow for tracking\cite{Wang2011}. The first point is the most reliable in terms of matching the position of a possibly salient point.

We filter such that a dense trajectory $x$ originating at $\pmb{o}_x = (o_x^{\hat x}, o_x^{\hat y})$ starting at frame $n$ is compared to the target's pose joint keypoints of the same frame $P_{n,m} = \{\pmb{p}_{n,m,1}, ..., \mathbf{p}_{n,m,j}, ..., \pmb{p}_{n,m,8}\}$. Trajectory $x$ is selected if $||\pmb{o}_x,\pmb{p}_{n,m,j}|| < R_{n,m,j}$ for any joint $j$. We also compare $x$ with poses in $P_{n-1}$ and $P_{n+1}$ using the same criterion. We found that comparing with 3 frames to adds robustness against inconsistent keypoint detections (common due to the frame-wise pose estimation). Using larger comparison windows tended to add increasingly more noise.

In our data, participants vary greatly in pixel size depending on their distance relative to the camera. To account for this, we scale $R_{n,j}$ according to the distance from the camera of the particular keypoint, ie. $R_{n,m,j} = R_j*S(\pmb{p}_{n,m,j})$, where $S$ computes a scaling factor that depends on the position of the point and $R_j$ are hyper-parameters.

We obtain the scaling factor $S$ via the camera-to-ground plane homography. We compute the transformation $A$ between the ground plane and the image plane using marks placed at known distances on the floor plane during data collection. This allows us to approximate a scale ratio between point $p_{n,j}$ and a reference point $p_r$ in the image plane as $||A(p_{n,j}), A(p_{n,j}+\Delta p)|| / ||A(p_r), A(p_r+\Delta p)||$ where $\Delta p$ is an arbitrarily small displacement vector.

\subsubsection{Extracting Fisher vector representation}

The chosen trajectories from different joints must be aggregated into a video-level representation. We make use of Fisher vectors \cite{Sanchez2013}, the state-of-the-art method for dense trajectory aggregation. Fisher vectors, and in particular their improved variant \cite{He2005} have been found to perform remarkably well in large scale datasets of daily activities like the recent Charades \cite{Sigurdsson2016, Sigurdsson2017a}.

Fisher vectors provide a compact feature representation from an arbitrary number of dense trajectories. Let $\pmb X = \{\textbf{x}_t \mid t=1, \dots, T\}$ be the set of $T$ dense trajectories of dimensionality $D = 426$ selected from keypoint regions and $u_{\pmb\lambda}$ be the probability density function with parameters $\pmb\lambda$. The fisher score is defined as the gradient of the log-likehood over $\pmb{X}$, with respect to the model parameters:

\begin{equation}
\pmb{{G}_{\lambda}^{X}} = \frac{1}{T} \nabla_{\pmb\lambda} \log u_{\pmb\lambda} (\pmb X)
\end{equation}

The Fisher vector is a normalized version of the Fisher score:

\begin{equation}
\mathcal{G}_{\pmb\lambda}^{\pmb{X}} = \pmb{L_{\lambda}G_{\lambda}^X}
\end{equation}

\noindent where normalization by $\pmb{L_{\lambda}}$ corresponds to whitening of the dimensions, where generative model can take the place of $u_{\lambda}$.

Normally a Gaussian mixture model (GMM) with $K$ components and diagonal covariance matrices is used as $u_{\lambda}$. The parameters $\pmb\lambda$ of a GMM are $\pmb\lambda=\{w_i, \pmb{\mu}_i, \pmb{\sigma}_i^2, i=1,\ldots,K\}$, where $w_i$, $\pmb{\mu}_i$ and $\pmb{\sigma}_i^2$ are the mixture weight, mean vector and diagonal of the covariance matrix of Gaussian $i$. However, only the mean and standard deviation are used because mixture weights add little additional information \cite{He2005}. Under the assumption of independence of local descriptors:

\begin{equation}
\pmb{G_{\lambda}^X} = \frac{1}{T} \sum_{t=1}^T \nabla_{\pmb\lambda} \log u_{\pmb\lambda} (\pmb x_t)
\end{equation}

Let $\gamma_t(i)$ be the soft assignment of descriptor $\pmb x_t$ to Gaussian $i$:

\begin{equation}
\gamma_t(i) = \frac{w_iu_i(\pmb x_t)}{\sum_{j=1}^K w_j u_i (\pmb x_t)}
\end{equation}

Calculation of the gradients leads to:

\begin{equation}
\mathcal{G}_{\mu, i}^{\pmb X} = \frac{1}{T\sqrt{w_i}} \sum_{t=1}^T \gamma_t(i) \left(\frac{\pmb x_t-\pmb\mu_i}{\pmb\sigma_i}\right)
\end{equation}

\begin{equation}
\mathcal{G}_{\sigma, i}^{\pmb X} = \frac{1}{T\sqrt{2w_i}} \sum_{t=1}^T \gamma_t(i) \left[ \frac{(\pmb x_t-\pmb\mu_i)^2}{\pmb\sigma_i^2}-1\right]
\end{equation}

\noindent where the division between vectors is term-by-term. The Fisher Vector aggregates all gradients into a vector of $2KD$ dimensions. For $K=256$ (used in previous work \cite{He2005}), this is a 218112-dimensional vector. 

Finally, the Fisher vectors are normalized by dividing by their L2 norm and then power-normalized with $f(z) = sign(z)\sqrt{|z|}$. These techniques have been shown to increase the ability of the Fisher Vector to detect subtle elements, and reduce the sparsity of Fisher vectors \cite{He2005} respectively.

A kernel on these gradients is defined as:
\begin{equation}
K(X,Y) = G_{\lambda}^{X'} F_{\lambda}^{-1} G_{\lambda}^{Y} = G_{\lambda}^{X'} L_{\lambda}'^{-1}  L_{\lambda}^{-1} G_{\lambda}^{Y}
\label{fisher_kernel}
\end{equation}

where $F_{\lambda}$ is symmetric and positive definite, and generally approximated such that normalization by $L_{\lambda}$ corresponds to a simple whitening of the dimensions.

Linear methods (traditionally linear SVM) are standard for classification of the FVs [30,38] because learning a linear classifier on the FVs is equivalent to learning a classifier using the Fisher kernel (kernel trick) and linear methods have delivered good results in previous work [33].

\subsection{Voice activity estimation from wearable acceleration }

\begin{figure}[h]
	\centering
	\includegraphics[width=0.5\textwidth]{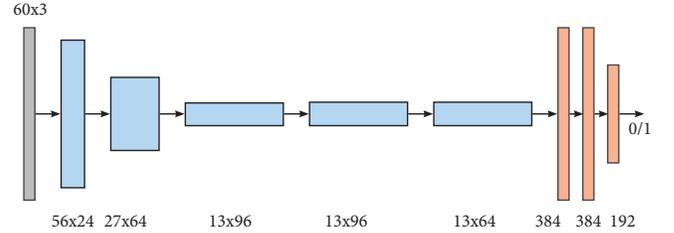}
	\caption{Architecture of the 1D-CNN used.}
	\label{fig:alex}
\end{figure}

Due to the good results obtained by deep methods in sensor-based activity recognition \cite{Wang2018a}, we use a one-dimensional CNN for acceleration-based detection. Just as previous work which makes use of relatively shallow CNNs for detecting actions from a single-accelerometer \cite{Chen2015, Gjoreski} we use a flattened version of the two-dimensional AlexNet \cite{Monien2007}, where we preserve the ratios between number of channels. Fig. \ref{fig:alex} shows the architecture used. Input data has 3 channels corresponding to axes X, Y and Z of the accelerometer. Filter sizes are 5 for the first convolutional layer and 3 for the rest of the layers, with unit padding. As with AlexNet, first, second and last layers are followed by a max-pooling layer with kernel size 3 and stride of 2.

The input to the CNN is pre-processed by subtracting the mean and dividing by the standard deviation, for each axis, to reduce the effect of gravity and device miscalibration.

\subsection{Multimodal fusion}

Both unimodal classifiers described above provide a posterior probability of the voice activity. We combine scores via late fusion to obtain multimodal scores. We make use of a Logistic Regression classifier without regularization, trained on both sets of scores, to obtain an intercept and weights for each classifier. We opt for score fusion because we expect acceleration scores to be complimentary (uncorrelated) to video in many cases when gestures or other vocalisation-associated body movements are hard to observe due to occlusion or orientation. While the acceleration signal encodes chest motion specifically, Fisher Vectors encode a mixture of visual and motion information from different body parts.

\section{Experiments and discussion}

\subsection{Data preparation and label aggregation}

Our pose extraction and tracking processes (section \ref{sec:pose_tracking}) results in variable-length pose tracks, for both datasets. Because these tracks vary wildly in length and to obtain a consistent sample length to use in evaluation, we split the obtained tracks (and accelerometer readings for our Free-standing dataset) in 3-second segments, each of which constitutes one data sample. Previous work has found windows of 3s to be maximally informative in speaking status detection tasks \cite{Gedik2017}. We labeled examples using a threshold on the fraction of positive VAD labels in the segment. Rather than using majority voting (0.5 threshold) we opted for a more aggressive threshold of 0.25 that would label most examples with speech activity as positives, which resulted in a more balanced dataset. Table \ref{tab:modalities} shows the label statistics.

\subsection{Training and hyper-parameter tuning}

In our Fisher vector pipeline, we extract dense trajectories with a length of $L=15$ and a sampling stride of $W=5$. These settings were found to be optimal in the original paper on dense trajectories \cite{Wang2011} and have been used as standard in more recent work \cite{Wang2013,Peng2016}. We train the GMM using a sample of 100000 trajectories, to which we apply Principal Component Analysis (PCA) preserving 95\% of the variance and whitening. We set $K=256$ components for the GMM and apply power normalization ($\alpha=0.5$) and L2 normalization to the Fisher vectors. All of these parameters and transformation were found to be optimal accross a variety of datasets in in previous work on best practices on training Fisher vector models \cite{He2005}. We classify Fisher vectors into positive and negative speaking status labels using a linear SVM with an L2 regularizer, following the same literature \cite{He2005}. Training was done via stochastic gradient descent (SGD). The optimal regularization parameter was found via 4-fold cross-validation.

We tuned the hyperparameters $R_j$ of our pose-aligned trajectories approach via experiments on a small set of data. Intuitively, these parameters determine the radius around each body keypoint from which trajectories are sampled. For simplicity, we considered $R_j = R \forall j$ (ie. the same radius is considered around every keypoint) and tested a set of 5 parameter settings (which we determined visually from data samples) on the held-out set via 4-fold cross-validation, which led us to a setting for $R$.

For the acceleration stream, we train the network using a binary cross-entropy loss and the Adam optimizer. Late fusion is performed by training a logistic regressor without regularization on the output scores of both modalities.

\subsection{Evaluation}

We evaluate all models via 10-fold cross-validation. The way in which the data is split is of relevance in our case. To avoid having significant dependencies between examples in the training and test sets, a person-level cross-validation split would be ideal. Due to the low number of participants, however, we opted for a split where every person-camera combination is considered one group, such that examples of the same person viewed from the same camera are always put together in either the train or the test set. 

We use the area under the ROC curve (AUC) as main evaluation metric as it quantifies the ability of the method to separate positive and negative labels in the output space, while being robust against dataset imbalance. We make use of Platt scaling to obtain probability scores from the SVM outputs.



\subsection{Impact of pose-based filtering on detection of speaking}\label{sec:filtering_results}

Previous work \cite{Cabrera-Quiros2018b} has shown that hands and arms produce informative trajectories for speaking status detection. We hypothesized that the selection of trajectories around skeleton keypoints will result in a more informative, less noisy set of trajectories that will be able to achieve greater generalization.

To test the improvement obtained by trajectory selection, we compare our method with a traditional bounding box approach. We obtained the bounding box for an example from the pose tracks by computing the bounding box that contains the subject's skeleton. For consistency, we computed bounding boxes for the upper body only, given the frequent occlusion of the lower body. A padding scaled using $A$ (see section \ref{sec:pose-aligned}) was added such that the subjects' upper body would be fully contained inside the bounding box.

We test two variants of our method. Table \ref{tab:results} shows the results, including the number of trajectories considered in each case. It indicates that our method improves over the baseline despite using only a subset of trajectories. In the first variant of our method (\textit{FV - UpperBody}) we extract trajectories around all upper body keypoints as presented in section \ref{sec:pose-aligned}. In the second (\textit{FV - HandsAndHead}) we select trajectories only around the head keypoint, and both wrist keypoints. This last method delivers the best results, while using on average only 34\% of the original trajectories. Both methods improve over the baseline (\textit{FV-Full}) by a significant margin. As an extra reference, we add a \textit{FV-Sampled} method which corresponds to random sampling of the trajectories with probability $p=0.34$, such that each example has on average the same number as selected by our method.

\begin{table}
	\caption{Results of 10-fold cross-validated experiments comparing our pose-based selection method with other methods based on dense trajectories. FV stands for Fisher Vectors.}
	\label{tab:results}
	\centering
	\begin{tabular}{lll}
		\toprule
		Method & Trajectories / example & AUC \\
		\midrule
		FV - Full & 1550.18 (1069.01) &  0.686 (0.015)\\
		FV - Sampled & 527.71 (345.93) & 0.678 (0.019)\\
		\hline
		FV - UpperBody & 619.67 (519.31)  &  0.713 (0.016)\\
		FV - HandsAndHead & 522.98 (458.82)  & 0.715 (0.020)\\
		\bottomrule
	\end{tabular}
\end{table}


\subsubsection{Role of cross-contamination}

\begin{figure*}[t!]
	\includegraphics[width=\textwidth]{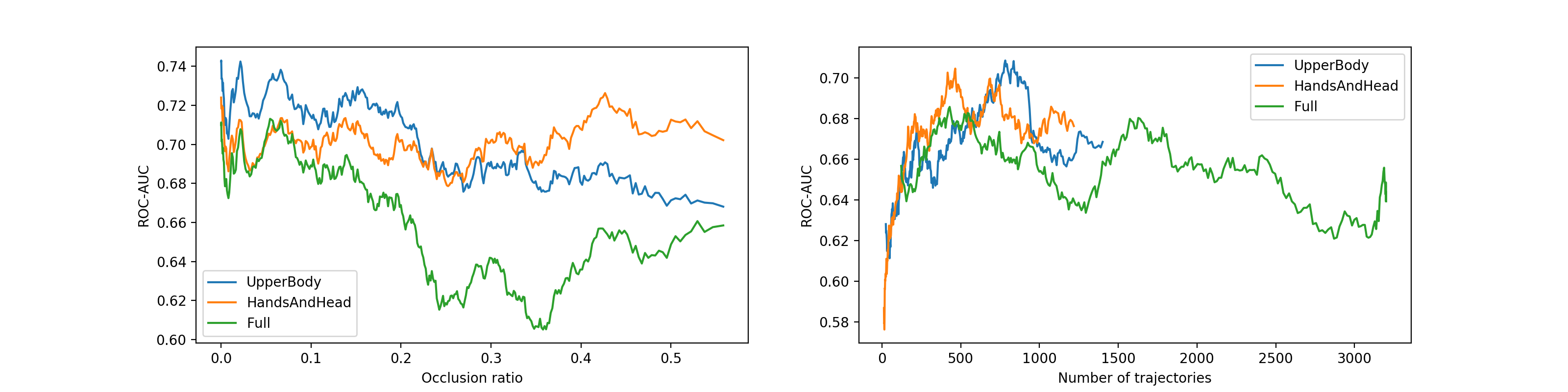}
	\caption{Left: AUC scores as a function of the cross-contamination score. Right: change in AUC scores with the number of trajectories in the example.}
	\label{fig:occlusion}
\end{figure*}

Since the initial motivation of our method is to avoid cross-contamination in a crowded setting, we further investigated its role in binary speaking status detection. We consider cross-contamination to include any case where the bodies of other people in the scene are visible within the bounding box of the target person. To this end, we gave every example a cross-contamination score. First, for a target person bounding box $B$ (section \ref{sec:filtering_results}) in frame $n$, we compute its cross-contamination score as:

\[
CC_B^n = \frac{\sum_i{Intersection(B, B_i^n)}}{Area(B)}
\]

\noindent where $B_i^n$ is the bounding box of pose $i$ in frame $n$. The score is the ratio between the intersection of the target pose's bounding box with all other detected poses' bounding boxes and the area of the target bounding box. To obtain an example-level score we simply take the median of its frame scores, in order to remove the effect of outliers due to differences in the estimated poses.

While this is not a perfect measure of cross-contamination, in our experiments we found this measure to consistently give high scores to segments where the target had was significantly occluded by another person, and decrease with less severe cases of cross-contamination.

Fig. \ref{fig:occlusion} shows the results of plotting the ROC AUC score as a function of the cross-contamination score. Here, it can clearly be seen that our method is specially stable regardless of cross-contamination level, while a bounding box is inevitably affected by it. The model considering only hands and head is also significantly more robust than the one considering all upper body keypoints. The number of trajectories in the example is also shown to have an influence, with examples with few trajectories being classified worse.

\begin{figure}
	\includegraphics[width=\columnwidth]{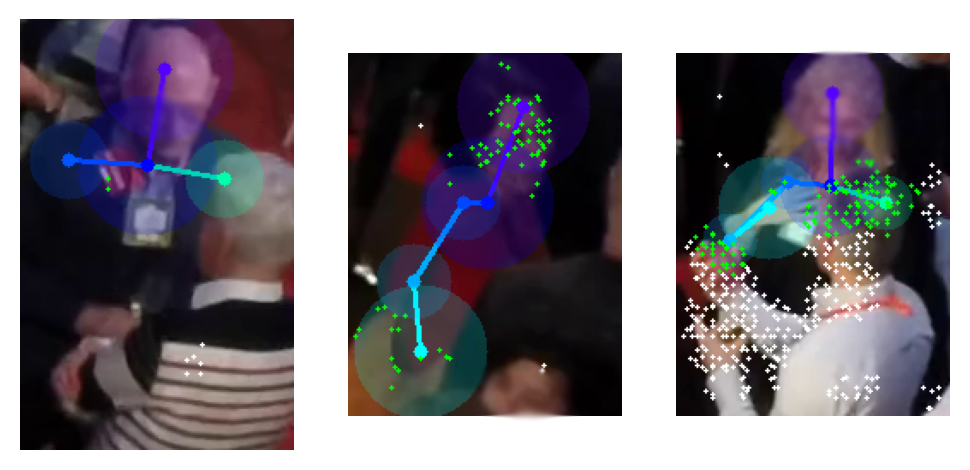}
	\caption{Some examples of cross-contamination. The target person whose speaking status is predicted is indicated by the presence of a skeleton. The initial xy-coordinates of trajectories within a bounding box are plotted for one frame. Points in green correspond to included trajectories and points in white correspond to trajectories that are discarded by our method. }
	\label{fig:high_occlusion}
\end{figure}

Fig. \ref{fig:high_occlusion} shows some examples of segments with high cross-contamination scores where the target subject's interlocutor occludes the subject's body. Points indicate the origin of a trajectory, with green points indicating trajectories selected by our method and white ones being discarded. In the third case our method avoids significant contamination from the target's interlocutor due to occlusion.

\subsubsection{Role of body parts and descriptors}

In this section, we evaluate the relative importance of different body parts in speaking status detection performance. In addition to our Free-standing dataset we evaluate on the RealVAD dataset since we wish to evaluate the importance of different body parts in a dataset with more consistent visibility of all body parts and a different (frontal) camera view angle. Because of the results of the previous section, indicating that the method works equally well with only hand and head keypoints, we focus only on these keypoints.

\begin{table*}
	\caption{AUC performance for 10-fold cross-validated speaking status detection, comparing the effect of using only information from hand or head keypoints. \textit{excl.} indicates that examples without the corresponding body part have been excluded from the computation.}
	\label{tab:body_parts}
	\centering
	\begin{tabular}{l|llllll}
		\toprule
		\multirow{2}{*}{Dataset} & \multirow{2}{*}{Body parts} & \multicolumn{5}{c}{Descriptors} \\
		 &  & Trajectory & HOG & HOF & MBH & All \\
		\midrule
		\multirow{6}{*}{FreeStanding} & Head & .6751 & .6958 & .6556 & .7018 & .7438\\
		& Head (excl.) & .6407 & .7056 & .6580 & .7138 & .7467 \\
		& Hands & .5835 & .5927 & .5974 & .6123 & .5865\\
		& Hands (excl.) & .5620 & .5288 & .5755 & .5842 & .6015\\
		& Hands\&Head & .6834 & .6965 & .6680 & .7163 & .7372\\
		& Hands\&Head (excl.) & .6648 & .7138 & .6717 & .7177 & .7484\\
		\midrule
		\multirow{3}{*}{RealVAD} & Head & .7980 & .7464 & .8115 & .8521 & .8017\\
		& Hands & .8230 & .7681 & .8334 & .8332 & .8475\\
		& Hands\&Head & .8650 & .7876 & .8537 & .8970 & .8743\\
		\bottomrule
	\end{tabular}
\end{table*}

Table \ref{tab:body_parts} shows the results of an ablation study where we remove features coming from either the hands or the head. We also reduce the descriptor set by only taking HOG, HOF or MBH descriptors for all trajectories. This effectively reduces the dimensionality of the Fisher vectors. We are most interested in the overall performance of the method when all features are used. Here it can be seen that the method using only head trajectories had in some cases better performance, although the differences between head-only and head-and-hands methods are small. The results show clearly that the hand movement information is less important, with a nearly 10\% difference in most cases.

The situation is different for the RealVAD dataset, where both hands and head are similarly informative. The higher scores when compared to the freestanding dataset can be explained by the setting, where people are sitting and relatively close to the camera. The classifier with only MBH features performed better than when all features were used due possibly to the influence of MBH features from the head. Inspection of the data revealed a possible reason for the fact that hands and head now had a similar influence: in RealVAD listeners tended to keep their hands on their laps (or otherwise completely still) most of the time, while the current speaker would very frequently gesture with their hands. Overall, however, it stands out that head movements are a strong predictor across both datasets.

We conclude therefore that the success of the filtering approach is driven mainly by information from head trajectories. This can be explained by the fact that in our freestanding dataset hands are more frequently occluded, especially behind the body depending on the person's orientation. To understand how occlusion of the hands affects the results, we also computed AUC scores excluding examples where the hands and/or head are not visible. These results are indicated with \textit{excl.} in Table \ref{tab:body_parts} and they show that the performance improvement is mild even when considering only examples where the body part is visible. This is for us strong indication that the head contains most of the information relevant to speaking status detection.

\subsubsection{Sensitivity analysis of parameters}

We perform a sensitivity analysis in an attempt to understand the importance of different hyper-parameters in model performance. We keep the dense trajectory hyper-parameters fixed following the standards of previous literature, which have found them to be close to optimal for different datasets. We focus on the filtering and Fisher vectors hyper-parameters. We start by experimenting with the GMM size $K$.

\begin{figure}
	\centering
	\includegraphics[width=0.8\columnwidth]{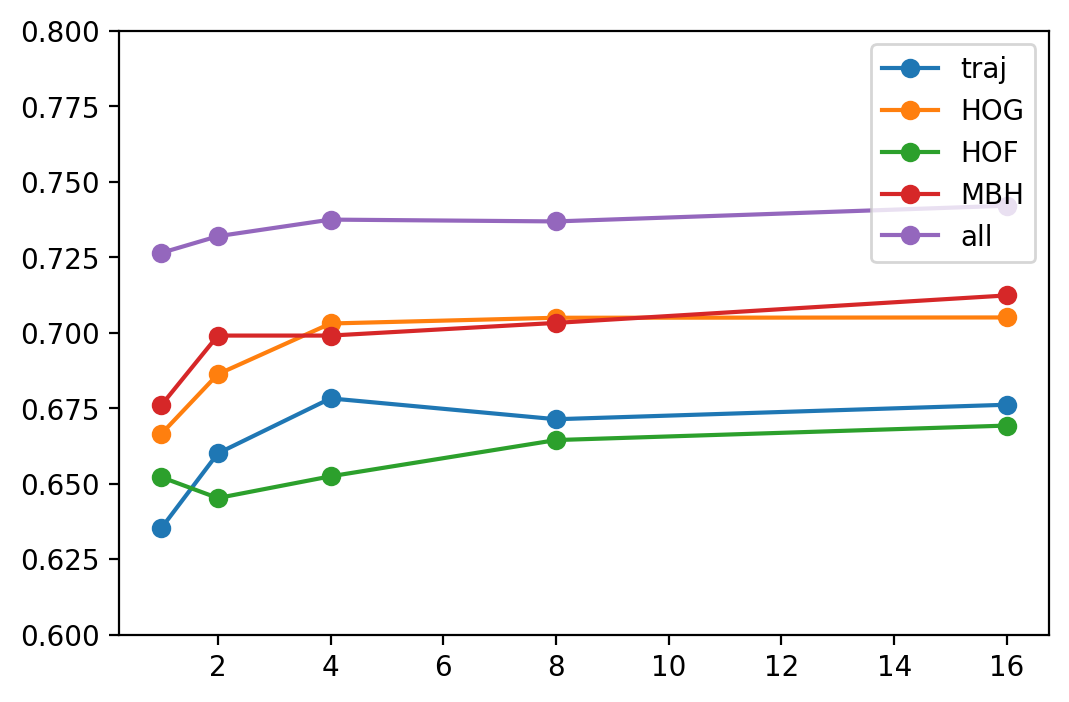}
	\caption{AUC scores in speaking status detection from video using different GMM sizes for Fisher vector extraction. Scores plateau for GMMs with 4 or more components.}
	\label{fig:gmm_size}
\end{figure}

Fig. \ref{fig:gmm_size} shows the AUC scores obtained for different number of components in the GMM. We show GMM sizes between 4 and 16 because this is where we could observe most variation. GMM sizes greater than 16 resulted in no significant increase in performance, indicating that in our case the method can be simplified further simply by using a GMM smaller than the standard 256. This indicates that features likely follow a distribution with few modes. This can be the case in our dataset due to its uniform setting, in contrast with the diversity of settings, backgrounds, subjects, and even video qualities present in action recognition datasets.

\subsection{Multimodal speaking status detection}

Table \ref{tab:accel_results} shows the results of our multimodal speaking status detection method (including the acceleration stream) compared with previous work. We compare with a baseline consisting on augmenting the acceleration signal with the magnitude and absolute value of each axis, followed by the computation of a power spectral density (PSD), binned into 8 logarithmically spaced bins. We follow the implementation detailed in previous work, including classification using a logistic regressor \cite{Gedik2017}. We did not compare with the personalized models proposed in such work due to the very low number of participants in our dataset. The logistic regressor hyper-parameter $C$ was tuned in a nested cross-validation loop.

\begin{table}
	\caption{Results of 10-fold cross-validated speaking status detection experiments comparing our 1-D CNN acceleration-based method with previous work.}
	\label{tab:accel_results}
	\centering
	\begin{tabular}{lll}
		\toprule
		Method & AUC \\
		\midrule
		PSD + Logistic Regression &  0.698 (0.025)\\
		1D-CNN &  0.738 (0.029)\\
		\bottomrule
	\end{tabular}
\end{table}

The results in Table \ref{tab:accel_results} indicate that our 1-D CNN clearly outperforms previous work. This is no surprise given that a CNN is able to learn complex features directly from the signal. 

Table \ref{tab:resultsmulti} shows the results of our multimodal approach with late fusion. These experiments are run only on the subset of the examples with both video and accelerometer data. Results suggest that both modalities indeed have a high degree of complementarity. 

\begin{table}
	\caption{Results of 10-fold cross-validated experiments on our multimodal fusion approach.}
	\label{tab:resultsmulti}
	\centering
	\begin{tabular}{lll}
		\toprule
		Method & AUC \\
		\midrule
		FV - HandsAndHead  & 0.720 (0.031) \\
		1-D CNN &  0.738 (0.029)\\
		Multimodal &  0.763 (0.027)\\
		\bottomrule
	\end{tabular}
\end{table}

\section{Limitations and future work}

Although we showed that automatic pose estimation methods are viable as person localization, it is also true that the extracted poses are not always reliable. The angle of the camera, illumination and occlusion all can significantly affect the quality of the pose estimation step. Our approach is not robust against mistakes of the pose estimator. This can potentially be improved through trajectory weighting instead of selection. In weighting, trajectories close to skeleton keypoints are weighted more than trajectories far from the keypoints. In this way, in cases where few of the visible body keypoints are found by the pose estimator, the method falls back to considering most trajectories with similar importance. This method, however, has the disadvantage of introducing some noise trajectories which could potentially offset its benefits. We are similarly interested in the utility of person instance segmentation methods in this step, but improvements in the quality and speed of such methods are necessary.

Our method requires the extraction of dense trajectories. While they are still an excellent option due to being quick to extract and easy to parallelize (both for extraction and processing with methods like Fisher Vectors) they have some clear drawbacks against neural-based approaches like their large and variable space requirements, and the inability to learn their parameters . Although not the primary goal of our work, we observed that one drawback of dense trajectories in this task in particular comes from the filtering of trajectories that are too short spatially. This filtering step is not ours, but is part of the base implementation of dense trajectories and is necessary to prevent an explosion in the number of trajectories extracted, but means that subtle cues related to speaking, like slight head or mouth movements are filtered out, leaving the method with no information in cases when there is no long-range movement of the limbs or body. For these reasons, we are interested in future work which combines these ideas with neural approaches, in an attempt to understand the importance of such subtle cues.

Regarding the acceleration modality, due to the low-dimensionality of the input, we think there is little benefit to be obtained from larger models. We believe future research on the use of this modality for speaking status detection should go in the direction of using more and higher frequency sensors. Research suggests that wrist and chest-worn sensors can be very informative of speaking status while remaining relatively unobtrusive and privacy-preserving. In regards to modality fusion, we are interested the exploration of smarter fusion approaches based on the observation that acceleration should have more influence in the prediction when the information available to the video stream is low.



\section{Conclusions}
In this work we presented a multimodal method combining wearable sensors with a pose-based video approach for speaking status detection in crowded settings, a challenging problem where the video modality can be severely affected by occlusion and cross-contamination. 

Using a dataset collected in the wild and annotated automatically for speaking status using high-quality voice recordings, we showed that pose detections from a state-of-the-art method are not only viable for person detection and tracking in a crowded scene, but that leveraging pose information in the action recognition stage improved performance while at the same time reducing the number of local features considered by the action recognition stage. This indicates a less noisy, more informative representation. The analysis of our method revealed that it is in cases of occlusion that our method is able to improve over the holistic approach, underscoring the advantage of using poses for person localization in a crowded scene.

Finally, the significantly  improved performance of the multimodal approach indicated that the video and acceleration modalities were complimentary.

We hope that these results will help inspire and adapt similar approaches that move towards improving the quality and speed with which machines are able to understand a crowded scene while reducing the human and computational time expenses.

\section*{Acknowledgements}
This research is supported by the Netherlands Organization for Scientific Research (NWO) under project number 639.022.606.

\ifCLASSOPTIONcaptionsoff
  \newpage
\fi



\bibliographystyle{IEEEtran}

\bibliography{library.bib}






\end{document}